\documentclass[sigconf,nonacm]{acmart}
\AtBeginDocument{%
  }

\usepackage{algorithm}
\usepackage[noend]{algorithmic}
\usepackage{colortbl}
\usepackage{multirow}
\usepackage{enumitem}
\usepackage{framed}
\usepackage{balance}

\newcommand{\gup}[1]{\tiny{\textcolor{teal}{\ #1$\uparrow$}}}   
\newcommand{\rdown}[1]{\tiny{\textcolor{red}{\ #1$\downarrow$}}} 
\newcommand{\gdown}[1]{\tiny{\textcolor{teal}{\ #1$\downarrow$}}} 
\newcommand{\rup}[1]{\tiny{\textcolor{red}{\ #1$\uparrow$}}}      

\newenvironment{insightbox}{%
  \def\FrameCommand##1{%
    \fcolorbox{yellow!50!black}{yellow!10!white}{##1}}%
  \MakeFramed{\advance\hsize-\width\FrameRestore}%
}{\endMakeFramed}

\begin{document}

\title{TRU: Targeted Reverse Update for Efficient Multimodal Recommendation Unlearning}

\author{Zhanting Zhou}
\orcid{0009-0006-9299-7355}
\affiliation{%
  \institution{University of Electronic Science and Technology of China}
  \city{Chengdu}
  \country{China}
}
\email{ztzhou@std.uestc.edu.cn}

\author{KaHou Tam}
\orcid{0000-0001-5816-6837}
\affiliation{%
  \institution{University of Macau}
  \city{Macau}
  \country{Macao}
}
\email{wo133565@gmail.com}

\author{Zeyu Ma}
\authornote{Corresponding author.}
\orcid{0000-0002-1846-8889}
\affiliation{%
  \institution{University of Electronic Science and Technology of China}
  \city{Chengdu}
  \country{China}
}
\email{cnzeyuma@163.com}

\author{Yang Yang}
\orcid{0000-0002-5070-4511}
\affiliation{%
  \institution{University of Electronic Science and Technology of China}
  \city{Chengdu}
  \country{China}
}
\email{yang.yang@uestc.edu.cn}

\author{Ziqiang Zheng}
\orcid{0000-0002-1477-6040}
\affiliation{%
  \institution{University of Electronic Science and Technology of China}
  \city{Chengdu}
  \country{China}
}
\email{zhengziqiang1@gmail.com}

\renewcommand{\shortauthors}{Zhanting Zhou, KaHou Tam, Zeyu Ma, Yang Yang, and Ziqiang Zheng}

\begin{abstract}
Multimodal recommendation systems (MRS) jointly model user-item interaction graphs and rich item content, but this tight coupling makes user data difficult to remove once learned. Approximate machine unlearning offers an efficient alternative to full retraining, yet current MRS unlearning applies reverse updates largely uniformly across model components. We show that this uniform treatment is misaligned with modern MRS: deleted-data influence is distributed unevenly across \textit{ranking behavior}, \textit{modality branches}, and \textit{model modules}. This non-uniformity gives rise to three bottlenecks in MRS unlearning: target-item persistence in the collaborative graph, modality imbalance across feature branches, and concentrated module-level sensitivity in the parameter space.
To address this mismatch, we propose \textbf{targeted reverse update} (TRU), a plug-and-play unlearning framework for MRS. Instead of applying a uniform global reversal, TRU performs three coordinated interventions across the model hierarchy: a ranking fusion gate to suppress residual target-item influence in ranking, branch-wise modality scaling to preserve retained multimodal representations, and capacity-aware parameter-group selection to localize reverse updates to deletion-sensitive modules. Across two backbones, three datasets, and three unlearning regimes, TRU achieves a stronger retain--forget trade-off than MMRecUn in most settings. In two challenging user-level cases, TRU also attains favorable operating points among all evaluated baselines. Security audits report the lowest MIA BalAcc and a tie for the lowest ASR among approximate methods in both audited settings, while wall-clock trajectories show earlier convergence to favorable retain--forget regions.
\end{abstract}

\ccsdesc[500]{Computing methodologies~Artificial intelligence}
\ccsdesc[500]{Security and privacy~Privacy protections}

\keywords{Machine Unlearning, Privacy Protection, Responsible Multimedia, Multimodal Recommendation}


\maketitle

\begin{center}
  \setlength{\fboxsep}{5pt}%
  \fbox{%
    \begin{minipage}{0.93\columnwidth}
      \centering\small\itshape
      Author Accepted Manuscript. Accepted for publication in the
      Proceedings of the 34th ACM International Conference on Multimedia (ACM
      MM '26). This author-created manuscript is not the ACM Version of
      Record. The definitive version is associated with DOI
      \url{https://doi.org/10.1145/3767308.3835081}.
    \end{minipage}%
  }
\end{center}

\section{Introduction}\label{introduction}

Multimodal recommendation systems (MRS) have become increasingly important in modern personalized platforms~\citep{mrs_survey_liu2023,fusion_survey25,huang2025survey}. By jointly modeling user-item interaction graphs and rich item content, such as images and text, they capture user preferences more accurately than interaction-only recommendation systems~\citep{zou2025realworld_recsys, mrs_survey_liu2023, mrs_survey_zhou2023}. 
However, this deep personalization also creates a serious privacy risk, as the tightly coupled multimodal signals in MRS may encode users' private and sensitive behaviors~\citep{ge2024trustworthy_recsys}. 
As data privacy regulations grant users the right to request deletion of their personal data \citep{gdpr2016art17, ccpa1798_105, pipl_official2021, pipl2021}, an MRS must remove both the data and its influence on the trained model.

\begin{figure}[t!]
  \centering
  \includegraphics[width=0.48\textwidth]{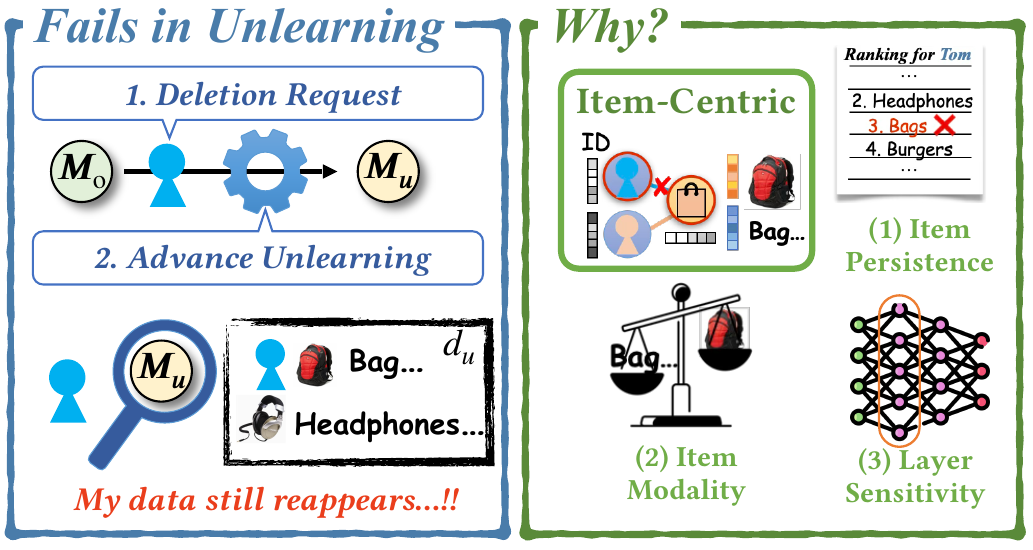}
  \vspace{-20pt}
  \caption{Conceptual overview of MRS unlearning. \textit{Left:} a deletion request triggers unlearning, but verification for the unlearned model fails. \textit{Right:} Ignoring the item-centric structure leads to these unlearning failures.}
  \Description{Two-panel schematic. The left panel traces a user-data deletion request through model unlearning to a failed verification outcome. The right panel attributes the failure to persistent item-side effects in a multimodal recommendation system.}
  \label{fig:preliminary}
  \vspace{-10pt}
\end{figure}

\begin{figure*}[!t]
  \centering
  \includegraphics[width=0.96\textwidth]{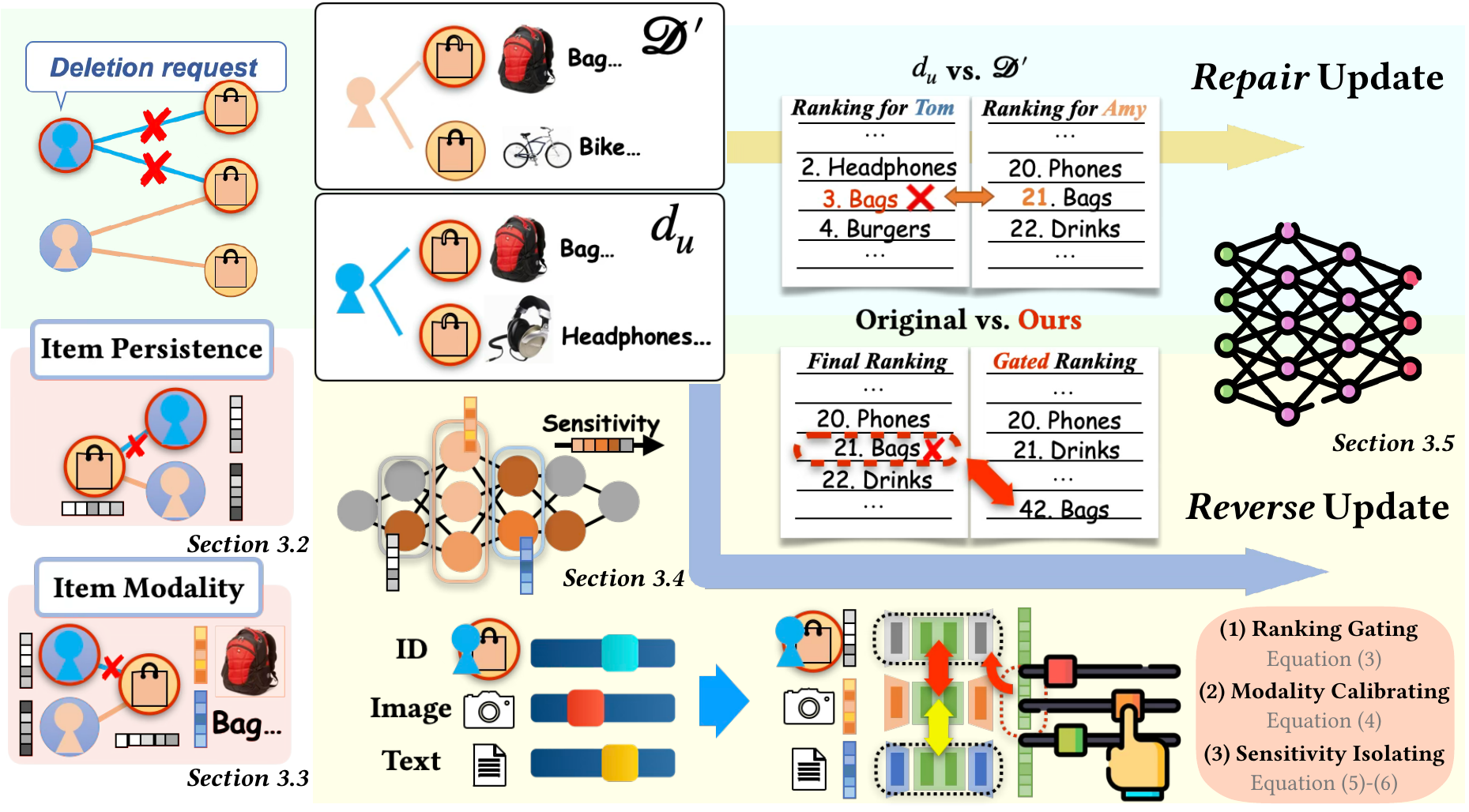}
  \vspace{-10pt}
    \caption{Overview of TRU. We diagnose three failure modes of uniform reverse unlearning in MRS: target-item effects persistence (Section~\ref{sec:ranking}), weak item-modality fusion (Section~\ref{sec:modality}), and module-level sensitivity (Section~\ref{sec:layer}). We map them to \textit{Ranking Gate}, \textit{Branch-wise Scaling}, and \textit{Parameter-Group Selection} in a unified reverse update (Section~\ref{sec:tru_unified}).}
  \Description{Pipeline diagram mapping three diagnosed failure modes—persistent target-item effects, imbalanced modality responses, and concentrated module-level sensitivity—to TRU's Ranking Gate, Branch-wise Scaling, and Parameter-Group Selection components before a unified reverse update.}
  \label{fig:method}
  \vspace{-10pt}
\end{figure*}

A straightforward way to satisfy such deletion requests is to retrain the entire model from scratch after each request. However, this strategy is computationally unsustainable in real-world systems, especially when deletion requests are frequent. As a result, \emph{machine unlearning} has emerged as a practical alternative for removing data influence without full retraining \citep{mu1,nguyen2025survey}. For practical unlearning in deployed MRS, it must satisfy three requirements simultaneously: \textbf{efficiency}, \textbf{forgetting fidelity}, and \textbf{retaining utility}. Specifically, it should process deletion requests without full retraining, thoroughly remove the influence of the target data, and preserve recommendation quality on the remaining data. The core challenge is that these objectives are inherently in tension~\citep{li2025recunlearn_survey, benchmark24}.

Existing studies address complementary parts of the unlearning problem, but none jointly handles the collaborative structure, multimodal branches, and ranked outputs of an MRS. Partition-based isolation approaches, \emph{e.g.}, RecEraser~\citep{recu} and UltraRE~\citep{ultrare}, provide structured deletion strategies without modeling the cross-modal coupling central to MRS. Relation-aware graph unlearning methods remove structural influence at the node or edge level, but do not explicitly model the heterogeneous multimodal components that shape item representations.
Conversely, multimodal unlearning methods such as MultiDelete~\citep{multidelete} erase paired multimodal samples while preserving cross-modal alignment, but do not target the collaborative graph structure or the resulting ranked recommendation behavior. MRS unlearning therefore requires a method that coordinates deletion across interaction structure, multimodal representations, and ranking outputs.


To bridge this gap, MMRecUn recently emerged as the first approximate unlearning framework tailored specifically for MRS~\citep{mmrecun}. By applying a reverse optimization step on forgotten data alongside a forward repair step on retained data, it avoids the cost of data partitioning~\citep{mmrecun}. Its reverse step, however, applies a largely \textbf{uniform reverse optimization} signal across the network. This uniform treatment is misaligned with the heterogeneous components of modern MRS. MGCN couples modalities through a shared modality-aware graph convolution, whereas MIG-GT uses modality-independent graph components and global transformers to accommodate modality-specific propagation scales and receptive fields~\citep{hu2025modality}. A single reverse signal therefore acts on components with different structures and response scales.

Recent studies~\citep{zhang2024modality,kim2024monet,hu2025modality} further substantiate this heterogeneity, showing that optimizing all modalities under a shared objective can induce modality imbalance and under-optimize less dominant branches. Our diagnostics show that approximate MRS unlearning is likewise not a uniform optimization problem: deleted-data influence appears as persistent target-item exposure in ranking, uneven responses across modality branches, and concentrated sensitivity across model modules. These observations identify three corresponding bottlenecks in MRS unlearning: \textbf{target-item persistence}, \textbf{modality imbalance}, and \textbf{module-level sensitivity}.


To address these challenges, we propose targeted reverse update (\textbf{TRU}), a plug-and-play framework for multimodal recommendation unlearning. Instead of applying a uniform global reverse update, TRU coordinates three interventions across ranking outputs, modality branches, and parameter groups: a prediction-aware \textit{ranking fusion gate} that suppresses residual target-item influence, \textit{branch-wise modality scaling} that calibrates reverse gradients for heterogeneous branches, and \textit{capacity-aware parameter-group selection} that localizes reverse updates to deletion-sensitive modules. TRU therefore determines \textbf{what} to suppress and \textbf{where} to apply the reverse update, respecting component-specific responsiveness.


Experiments across two representative backbones, three datasets, and three unlearning regimes show that TRU yields a stronger retain--forget trade-off than MMRecUn in most settings. In two challenging user-level settings, TRU also attains favorable operating points against UltraRE, MultiDelete, ScaleGUN, and MMRecUn. The security audits report the lowest MIA BalAcc and tied-lowest ASR among approximate methods in both evaluated cases, while the wall-clock trajectories show that TRU reaches favorable retain--forget regions earlier. Together with the cross-level diagnostics, these results support a clear design principle: effective MRS unlearning requires targeted control over ranking outputs, modality branches, and sensitive parameter groups. Our contributions are summarized as follows:
\begin{itemize}[leftmargin=*]
    \item \textbf{Cross-level diagnosis}. We show that deleted-data influence in MRS unlearning is uneven across ranking behavior, modality branches, and model modules, revealing target-item persistence, modality imbalance, and module-level sensitivity.
    \item \textbf{Targeted reverse update framework}. We present TRU, a plug-and-play framework that maps these three bottlenecks to a ranking gate, branch-wise modality scaling, and capacity-aware selection of sensitive parameter groups without changing the backbone or inference procedure.
    \item \textbf{Strong retain--forget trade-offs}. TRU improves on MMRecUn in most backbone--dataset--regime combinations and attains favorable trade-offs among broader baselines in the two challenging user-level settings; security and wall-clock audits provide complementary evidence beyond ranking metrics.
\end{itemize}

\section{Related Work}
\label{sec:related}

\subsection{Heterogeneity in MRS Architectures}
\label{sec:related_mr}

Multimodal Recommendation Systems (MRS) deeply couple collaborative interaction graphs with rich item-side content \citep{mgcn,hu2025modality} (\textit{e.g.}, images and text) to capture fine-grained user preferences \citep{mrs_survey_zhou2023}. While early approaches relied on simple feature concatenation \citep{mrs_survey_liu2023}, recent surveys indicate that state-of-the-art backbones have evolved into highly structured, heterogeneous architectures \citep{zou2025realworld_recsys,mrs_survey_liu2023,huang2025survey}. 
Specifically, graph-based models like MMGCN \citep{mmgcn}, GRCN \citep{grcn}, and MGCN \citep{mgcn} employ modality-aware convolutions and multi-view graphs to govern information propagation. 
Building upon these modality-aware designs, more advanced architectures push this structural decoupling even further. 
For instance, models like MIG-GT \citep{hu2025modality} use modality-specific receptive fields and tailored graph neighborhoods rather than a single shared topology.
Consequently, optimizing these decoupled branches via a single shared objective can lead to modality imbalance \citep{zhang2024modality}.
Adjacent federated-unlearning studies highlight analogous heterogeneity issues: domain-aware update allocation and verification require explicit treatment \citep{tam2025fudws,tam2025federated_domain_unlearning}.
From an unlearning perspective, applying uniform reverse updates across these decoupled architectures can affect retained multimodal branches and fusion pathways unevenly.
This observation motivates \textbf{targeted interventions} that account for architectural heterogeneity.

\subsection{Illusion of Uniformity in Unlearning}
\label{sec:related_mu}

While machine unlearning aims to efficiently erase specific data influences, its execution in recommendation is complicated by collaborative graph effects \citep{li2025recunlearn_survey}, where removing a single user or item can affect the representations of retained nodes through network propagation.
Existing literature largely bifurcates into partition-and-retrain frameworks (\textit{e.g.}, RecEraser \citep{recu}, UltraRE \citep{ultrare}) and approximate parameter updates (\textit{e.g.}, SCIF \citep{recu2}, IFRU \citep{ifru23}). 
Parallel efforts in graph unlearning, like GNNDelete \citep{gnndelete2023} and ScaleGUN \citep{scalegun_yi2024}, have advanced structural erasure by explicitly modeling node and edge removals. 
These graph-centric paradigms focus on structural removal without explicitly modeling heterogeneous multimodal components.
When applied to an MRS, they therefore do not directly address the joint effects of collaborative topology and multimodal branches.
Conversely, multimodal unlearning methods such as MultiDelete \citep{multidelete} address cross-modal decoupling but do not explicitly address collaborative ranking objectives.

To bridge this specific gap, MMRecUn \citep{mmrecun} recently emerged as the first approximate unlearning framework tailored for MRS, coupling a reverse BPR update \citep{bpr} on the forgotten data with a forward repair step. 
This framework uses a largely \textit{uniform} reverse optimization signal across the backbone.
Because modality branches can have distinct learning dynamics and representation spaces, we examine whether an identical reverse penalty can over-correct some modality-specific features while under-correcting others.
Such uneven responses can leave residual effects across modality branches and ranked outputs.
This motivates \textbf{targeted interventions} that calibrate reverse updates according to the distinct sensitivities of the modality branches during reverse optimization.

\section{Methodology} \label{sec:design}
We first define the problem formulation and explain the motivation of designing our TRU in Section~\ref{sec:setup}, where the framework overview is shown in Figure~\ref{fig:method}. 

\subsection{Problem Formulation}
\label{sec:setup}
Let $\mathcal{D}$ denote original training interactions, and let $d_u \subset \mathcal{D}$ be the deletion set specified by a \textit{user-}, \textit{item-}, or \textit{interaction-}level request. We refer to the process of data \textbf{deletion} as \textbf{unlearning}, which means that it forces the original model to \textbf{forget} the requested private data while preserving performance on \textbf{the retained} data:
\begin{equation}
\mathcal{D'} = \mathcal{D} \setminus d_u,
\label{eq:remain_data}
\end{equation}
and the goal of approximate MRS unlearning is to obtain an updated model whose behavior approaches the model retrained entirely on $\mathcal{D}'$, while avoiding the prohibitive cost of full retraining \citep{mu1}.

In generic machine unlearning~\citep{mu1,mu2,reverse}, a deletion request is often specified at the level of an individual training record, but the learned influence can still propagate through shared model parameters. In MRS, this challenge is further amplified because the influence of any target user, item, or interaction is \textbf{inherently entangled} with others. Concretely, in MRS, the influence propagates jointly through both the collaborative user–item graph and the deeply coupled modality branches, requiring the model to \textit{precisely remove the targeted effect} while \textit{preserving the shared representations} essential for the retained data.



Following the existing approximate MRS unlearning approach MMRecUn~\citep{mmrecun}, we adopt a \textbf{reverse-repair} unlearning protocol. Specifically, at each epoch, the model first updates on $d_u$ in the reverse direction of its gradient (\textit{i.e.}, gradient ascent) to remove its influence, and then performs a repair update on $\mathcal{D}'$ to recover performance on the retained data after deletion:
\begin{equation}
\begin{split}
\theta &\leftarrow \theta + \eta \alpha \nabla_{\theta}\mathcal{L}_{\mathrm{MRS}}(d_u),\\
\theta &\leftarrow \theta - \eta (1-\alpha)\nabla_{\theta}\mathcal{L}_{\mathrm{MRS}}(\mathcal{D'}),
\end{split}
\label{eq:reverse_repair}
\end{equation}
where $\eta$ is the learning rate and $\alpha$ controls the protocol trade-off. 

In this paper, we use item-side modalities to denote the three item-level signal sources handled by separate branches in the backbone. Our empirical analysis reveals that the key bottleneck in MRS unlearning lies in its \textbf{item-centric} nature: target items are not fully removed from ranking outcomes because of their collaborative dependencies with others, and their influences are unevenly propagated across different item modalities. To overcome this challenge, our TRU discards uniform global penalties and instead applies \textbf{targeted reverse interventions} that selectively counteract item-side effects while maintaining the structural coherence of the remaining collaborative graph. Specifically, we propose three key strategies: \textit{suppressing persistent target-item effects} (Section~\ref{sec:ranking}), \textit{calibrating modality-specific gradients} (Section~\ref{sec:modality}), and \textit{isolating sensitive parameter groups} (Section~\ref{sec:layer}) for precise and robust multimodal recommendation unlearning across heterogeneous backbones.

\subsection{Suppressing Persistent Target-Item Effects} \label{sec:ranking}

\begin{figure}[t!]
  \centering
  \includegraphics[width=0.48\textwidth]{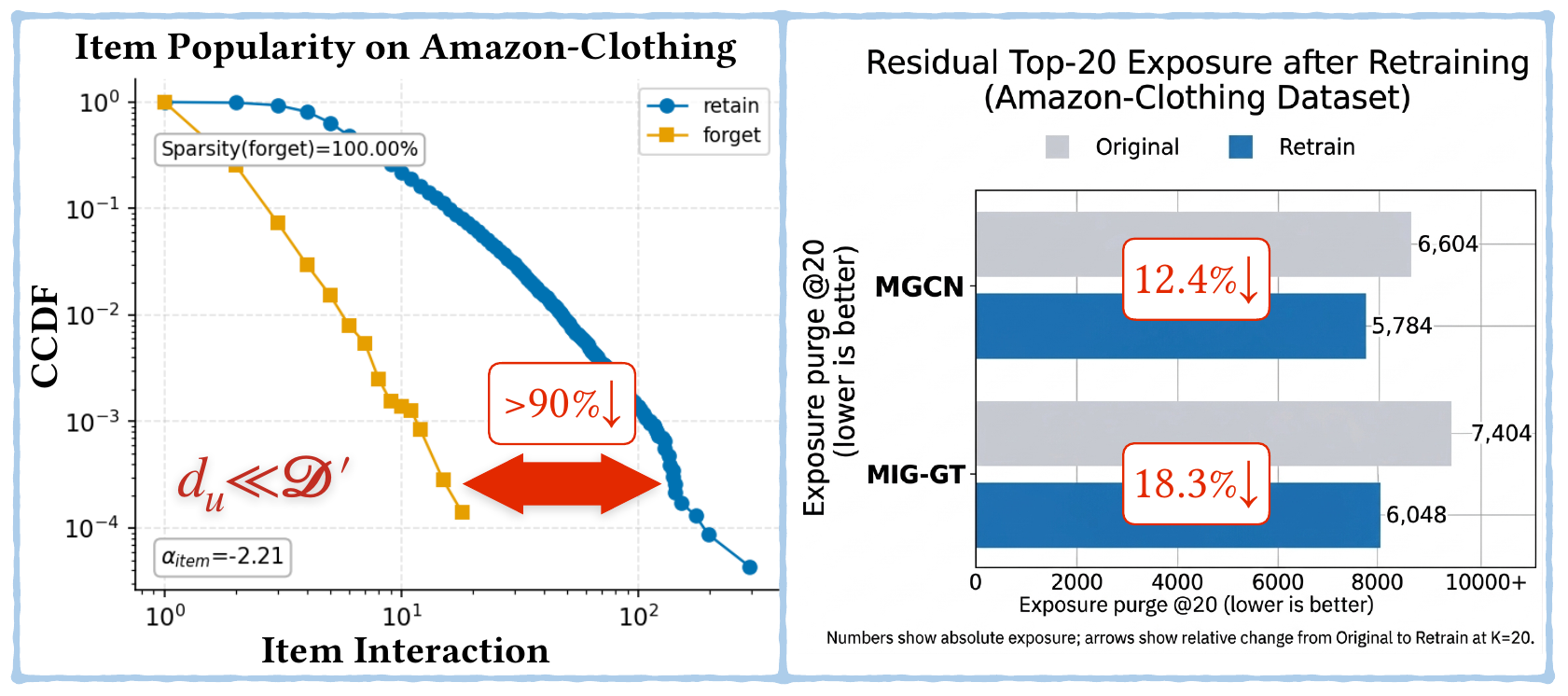}
  \vspace{-20pt}
  \caption{Item persistence on Amazon-Clothing. Left: the forget set is much sparser than the retain set in item popularity. Right: target-item exposure in the Top-20 remains non-zero even after retraining.}
  \Description{Two plots for Amazon-Clothing. The left compares target-item popularity in the forget and retain sets and shows a large sparsity gap. The right compares Top-20 target-item exposure before and after retraining and shows that exposure remains non-zero.}
  \label{fig:ranking_evidence}
  \vspace{-10pt}
\end{figure}

We empirically identify a persistent failure mode in MRS unlearning, where target items linked to deleted users continue to appear in the final recommendation lists due to lingering collaborative dependencies. To support this finding, we present Figure~\ref{fig:ranking_evidence}, which visualizes how the exposure of these items changes before and after data deletion. Even after a \textbf{full retraining} on the Amazon-Clothing dataset, the target items remain visible in the Top-20 rankings: exposure decreases from 6,604 to 5,784 (12.4\%) for MGCN~\citep{mgcn} and from 7,404 to 6,048 (18.3\%) for MIG-GT~\citep{hu2025modality}. We attribute the persistent target-item effects to the collaborative user-item graph. As illustrated in the left panel of Figure~\ref{fig:ranking_evidence}, the deleted interactions associated with the deleted users ($d_u$) represent a minor fraction (less than 10\%) of the overall engagement for most items, leaving a sparsity gap exceeding 90\% when compared to the retained data ($\mathcal{D}'$). This imbalance explains why deleting user edges fails to suppress item exposure: the residual popularity of those items is largely reinforced by other users who remain in the MRS.
\begin{insightbox}
    \begin{enumerate}[label={(\arabic*)}, leftmargin=*]
    \item \textbf{Collaborative popularity:} Shared connections with retained users keep target items visible, so simple edge deletion cannot remove them from ranked outputs.
    \item \textbf{Output suppression:} Directly penalizing the item's ranking score is necessary to counter its residual popularity.
    \end{enumerate}
\end{insightbox}





\subsection{Calibrating Modality-Specific Gradients} \label{sec:modality}

\begin{figure}[t!]
  \centering
  \includegraphics[width=0.48\textwidth]{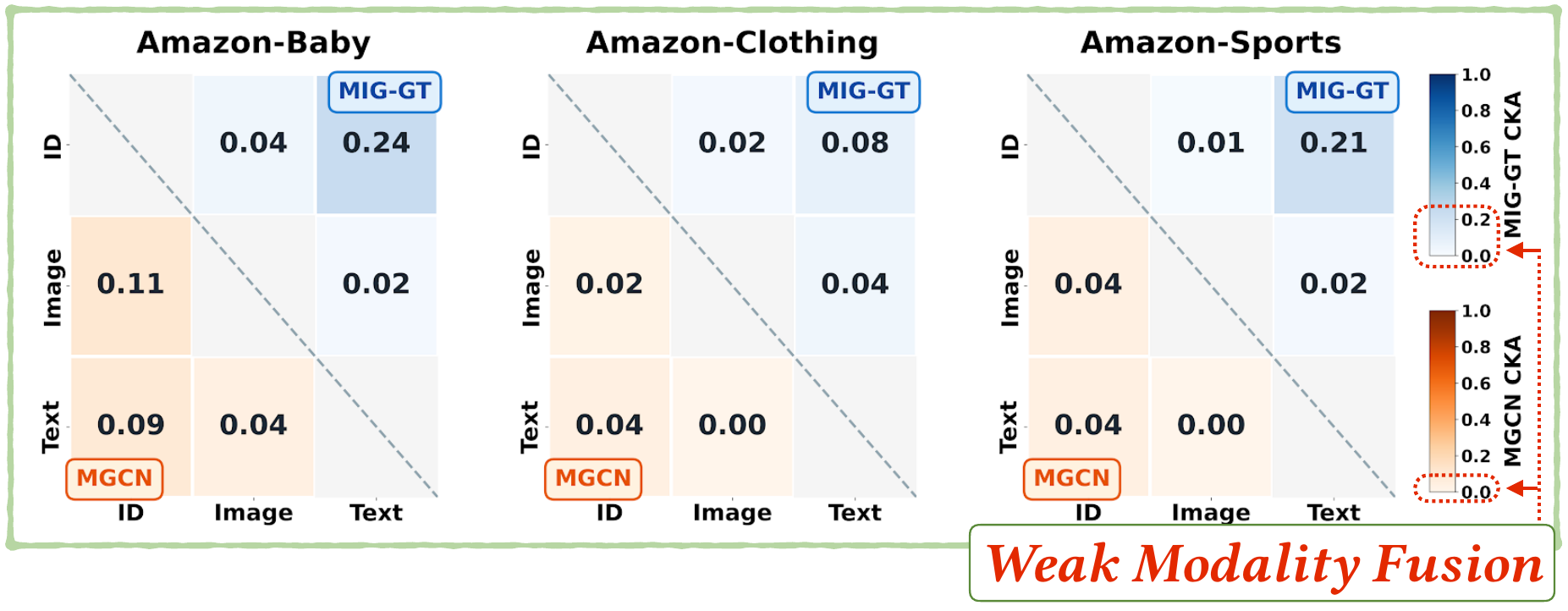}
  \vspace{-20pt}
  \caption{Item modality imbalance. Lower-left / upper-right: cross-modal alignment in MGCN / MIG-GT. Off-diagonal similarities remain weak in both backbones.}
  \Description{Two cross-modal similarity heat maps for MGCN and MIG-GT. Diagonal cells represent within-modality similarity, while the substantially smaller off-diagonal cells show weak alignment among ID, image, and text representations.}
  \label{fig:modality_evidence}
  \vspace{-10pt}
\end{figure}

Besides ranking persistence, we empirically identify a further bottleneck: item modalities absorb unlearning signals unevenly. As revealed in Figure~\ref{fig:method}, applying the same reverse penalty across the \textbf{ID}, \textbf{image}, and \textbf{text} branches leads to a pronounced imbalance, where we provide direct quantitative evidence in Figure~\ref{fig:modality_evidence}. We adopt centered kernel alignment (CKA)~\citep{Kornblith2019CKA,Davari2022CKAReliability} to demonstrate that the representations across these three branches are barely aligned. Specifically, both MGCN and MIG-GT exhibit very low cross-modal similarities, with most values falling below $0.1$.

The weak cross-modal alignment is not merely an artifact but a structural consequence of the decoupled architectures discussed in Section~\ref{sec:related_mr}. By design, modern MRS isolates propagation pathways to capture modality-specific preferences \citep{mgcn,hu2025modality,zhang2024modality}, thereby operating in distinct and weakly aligned representation spaces. Consequently, applying a uniform reverse optimization step across such heterogeneous branches is inherently unbalanced. Fragile modality branches tend to be over-corrected, degrading the representations of retained items, while dominant branches remain under-corrected, thus leaving traces of the deleted items partially preserved.
\begin{insightbox}
    \begin{enumerate}[label={(\arabic*)}, leftmargin=*]
        \item \textbf{Uneven forgetting:} Text, image, and ID branches process features independently, so they respond unequally to the same reverse signal during unlearning.
        \item \textbf{Modality-specific calibration:} The unlearning process must adaptively modulate gradient magnitudes for each modality to limit degradation of retained representations.
    \end{enumerate}
\end{insightbox}

\subsection{Isolating Sensitive Layers} \label{sec:layer}

\begin{figure}[t!]
  \centering
  \includegraphics[width=0.48\textwidth]{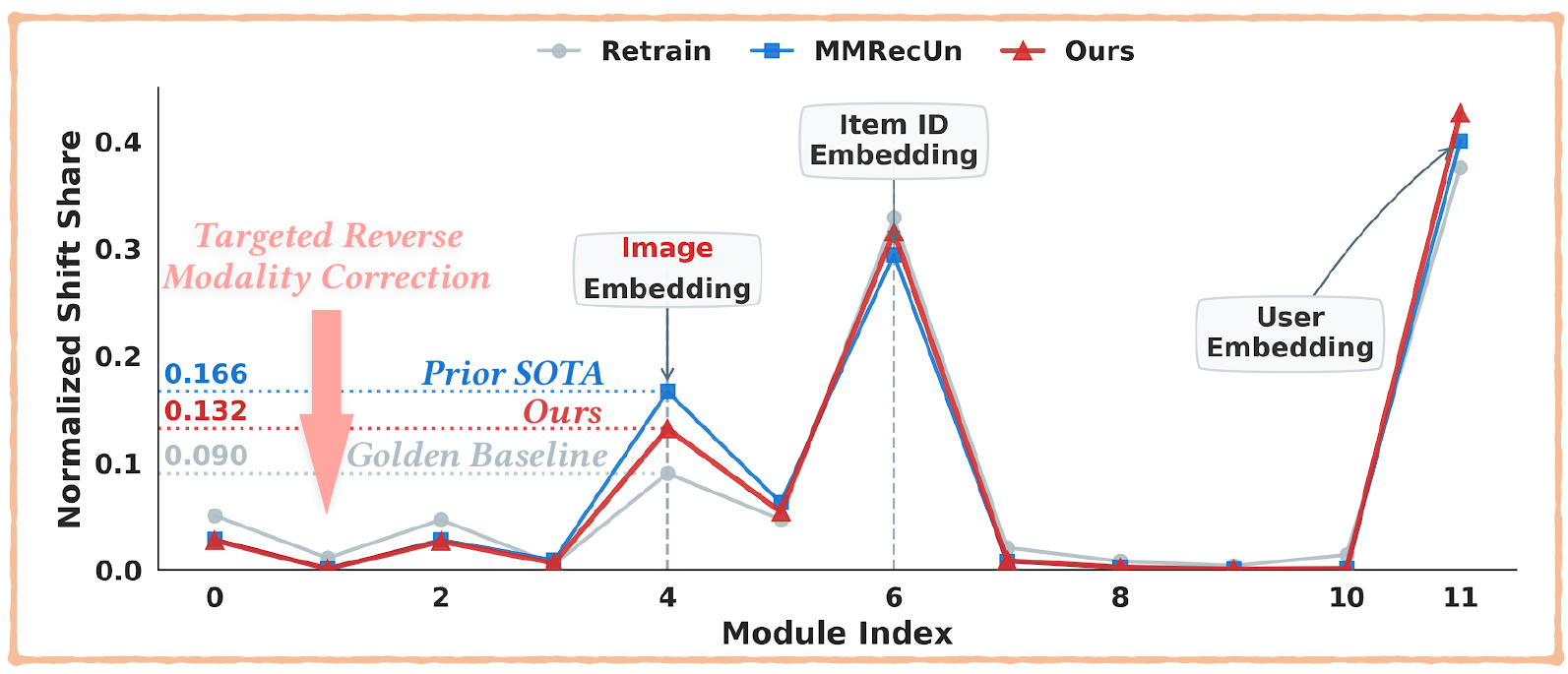}
  \vspace{-20pt}
  \caption{Layer sensitivity mismatch. MMRecUn over-shifts early item embedding modules relative to retraining, while TRU stays closer to the retraining profile.}
  \Description{Layer-wise parameter-shift comparison among retraining, MMRecUn, and TRU. MMRecUn shows larger deviations in early item-embedding modules, whereas the TRU profile is closer to retraining across the displayed modules.}
  \label{fig:layer_evidence}
  \vspace{-10pt}
\end{figure}

Building on the asymmetric behaviors observed in item rankings and modalities, we hypothesize that the unlearning signal propagates unevenly across the model's parameter space. To validate our hypothesis, we present Figure~\ref{fig:layer_evidence} to quantify the layer-wise imbalance in parameter update, showing that conventional methods (\emph{e.g.}, MMRecUn) excessively shift early embedding modules under uniform reverse update, whereas our TRU isolates interventions to achieve a balanced correction across layers. For instance, at the \texttt{Image Embedding} layer, MMRecUn's normalized parameter shift spikes to $0.166$, deviating severely from the exact retraining baseline ($0.089$). In contrast, our TRU effectively mitigates this structural over-reaction. By isolating the interventions, TRU reduces the shift to $0.132$, successfully pulling the parameter trajectory much closer to the golden retraining profile. These results quantitatively prove that a safe unlearning process must surgically act on sensitive modules rather than blindly updating the entire backbone, since the deletion sensitivity is concentrated in specific network modules.
\begin{insightbox}
    \begin{enumerate}[label={(\arabic*)}, leftmargin=*]
        \item \textbf{Layer-wise sensitivity:} Deletion effects concentrate in specific modules, especially early embedding layers.
        \item \textbf{Targeted isolation:} Unlearning should confine reverse updates to these layers so retained representations remain stable. 
    \end{enumerate}
\end{insightbox}

\subsection{Unified Targeted Reverse Update}
\label{sec:tru_unified}

To address the three unlearning bottlenecks identified above, we propose the targeted reverse update (TRU) framework. Building on our systematic analyses in Section~\ref{sec:ranking}–\ref{sec:layer}, TRU explicitly aligns unlearning signals with the underlying architectural heterogeneity. More crucially, TRU functions as a \textbf{plug-and-play} framework: it reformulates reverse optimization without changing the backbone, retained objective, or inference procedure.

\textbf{Mitigating Ranking Persistence with a Fusion Gate}. To mitigate the collaborative inertia that maintains the visibility of deleted items (Section~\ref{sec:ranking}), we apply an $\ell_1$ penalty to the selected gate parameters during the reverse step:
\begin{equation}
\mathcal{L}_{\mathrm{rev+gate}}(b_f)
=
\mathcal{L}_{\mathrm{swap}}(b_f)
+
\lambda_{\mathrm{gate}}
\sum_{\vartheta\in\mathcal{G}_{\mathrm{gate}}}
\|\vartheta\|_1,
\label{eq:ranking_gate}
\end{equation}
where $\mathcal{L}_{\mathrm{swap}}$ is the base ranking loss evaluated after swapping the positive and negative items in a forget batch $b_f$, and $\mathcal{G}_{\mathrm{gate}}$ is the selected set of gate-parameter tensors. The reverse step minimizes this objective, so the positive $\ell_1$ term encourages sparsity among the selected gates.

\textbf{Mitigating Modality Imbalance with Branch-Wise Scaling.}
As established in Section~\ref{sec:modality}, distinct item modalities react asymmetrically to identical reverse perturbations. For each modality branch $m \in \{\text{ID, image, text}\}$, we estimate retain- and forget-side gradient energies as
\begin{equation}
\begin{aligned}
R_m &=
\frac{1}{M_R}
\sum_{b_r\in\mathcal{B}_R}
\sum_{\vartheta\in\mathcal{G}_m}
\left\|\nabla_{\vartheta}\mathcal{L}_{\mathrm{MRS}}(b_r)\right\|_2,\\
F_m &=
\frac{1}{M_F}
\sum_{b_f\in\mathcal{B}_F}
\sum_{\vartheta\in\mathcal{G}_m}
\left\|\nabla_{\vartheta}\mathcal{L}_{\mathrm{swap}}(b_f)\right\|_2,
\end{aligned}
\label{eq:branch_energy}
\end{equation}
where $\mathcal{B}_R$ and $\mathcal{B}_F$ are sampled retain and forget minibatches, $M_R$ and $M_F$ are their counts, and $\mathcal{G}_m$ contains the parameter tensors assigned to branch $m$. Each tensor's gradient norm is computed before summing within a branch and averaging across batches; $F_m$ excludes the gate penalty. The reverse-gradient scale is
\begin{equation}
\gamma_m
=
\max\!\left(0.50,\;
\min\!\left(1,\;
\frac{\max(10^{-12},R_m)}
{\max(10^{-12},F_m)}\right)\right),
\label{eq:modality_scale}
\end{equation}
and is applied directly to the reverse gradient. Thus, the reverse gradient is attenuated to a factor between $0.50$ and $1$ only when $F_m>R_m$; otherwise, it retains full strength.

\begin{algorithm}[t]
\caption{\textsc{TRU}: Targeted Reverse Update}
\label{alg:tru}
\begin{algorithmic}[1]
\REQUIRE Backbone $B_{\theta}$; base loss $\mathcal{L}_{\mathrm{MRS}}$; retain/forget data $\mathcal{D}'$, $d_u$; top proportion $p$; min capacity $\tau_{\min}$
\FOR{each epoch}
    \STATE Estimate parameter-group sensitivities $\{e_k^{\mathrm{rev}}\}$ (Eq.~\ref{eq:layer_sensitivity}) and form mask $\{z_k\}$ (Eq.~\ref{eq:layer_mask})
    \STATE Estimate branch-wise scalers $\{\gamma_m\}$ (Eqs.~\ref{eq:branch_energy}--\ref{eq:modality_scale})
    \STATE Install hooks that mask and scale reverse gradients
    \FOR{forget batch $b_f \sim d_u$}
        \STATE Minimize $\mathcal{L}_{\mathrm{rev+gate}}(b_f)$ using Eq.~\ref{eq:tru_reverse_step}
    \ENDFOR
    \STATE Remove the reverse-gradient hooks
    \FOR{retained batch $b_r \sim \mathcal{D}'$}
        \STATE Minimize the original retain objective $\mathcal{L}_{\mathrm{MRS}}(b_r)$
    \ENDFOR
\ENDFOR
\RETURN updated backbone $B_{\theta}$
\end{algorithmic}
\end{algorithm}

\textbf{Capacity-Aware Parameter-Group Isolation.} To prevent the parameter over-shifting observed in Section~\ref{sec:layer}, we restrict the reverse update to selected parameter groups (modules). For group $k$, let $\mathcal{G}_k$ denote its parameter tensors. We compute
\begin{equation}
e_k^{\mathrm{rev}}
=
\frac{1}{M_F}
\sum_{b_f\in\mathcal{B}_F}
\sum_{\vartheta\in\mathcal{G}_k}
\left\|
\nabla_{\vartheta}\mathcal{L}_{\mathrm{swap}}(b_f)
\right\|_2.
\label{eq:layer_sensitivity}
\end{equation}
We sort the groups by decreasing sensitivity. The selected set $\mathcal{K}_{\mathrm{sel}}$ starts with the top $\lceil p|\mathcal{K}|\rceil$ groups and expands until it covers at least a fraction $\tau_{\min}$ of the model's named parameter tensors. The binary mask is
\begin{equation}
z_k
=
\mathbf{1}\!\left[k\in\mathcal{K}_{\mathrm{sel}}\right],
\label{eq:layer_mask}
\end{equation}
which confines the reverse update to the selected groups.

\begin{table*}[t!]
\centering
\caption{Performance comparison between the prior SOTA MRS unlearning method (MMRecUn) and Ours across diverse deletion regimes. TRU consistently pushes both Retain and Forget metrics closer to the exact retraining frontier, proving that its superiority stems from a fundamentally better trade-off control rather than isolated metric manipulation. The background colors denote the magnitude of performance improvement achieved by our method over the MMRecUn baseline. Darker/lighter colors represent larger/smaller performance differences (best performance is highlighted in bold). All ranking metrics are reported to four decimal places. Entries shown as 0 denote values below 0.00005 after rounding, not exact zero.\label{tab:comparison}}
\vspace{-0.15in}
\definecolor{RetainColor}{HTML}{4A90E2}
\definecolor{ForgetColor}{HTML}{F5A623}
\resizebox{\textwidth}{!}{
\begin{tabular}{c c l | ccc | ccc | ccc | ccc}
\toprule
\multirow{2}{*}{\textbf{Backbone}} & \multirow{2}{*}{\textbf{Dataset}} & \multirow{2}{*}{\textbf{Method}} & \multicolumn{3}{c|}{\textbf{RETAIN Recall@20 $\uparrow$}} & \multicolumn{3}{c|}{\textbf{RETAIN NDCG@20 $\uparrow$}} & \multicolumn{3}{c|}{\textbf{FORGET Recall@20 $\downarrow$}} & \multicolumn{3}{c}{\textbf{FORGET NDCG@20 $\downarrow$}} \\
\cmidrule(lr){4-6} \cmidrule(lr){7-9} \cmidrule(lr){10-12} \cmidrule(lr){13-15}
& & & User & Item & Inter. & User & Item & Inter. & User & Item & Inter. & User & Item & Inter. \\
\midrule
\multirow{9}{*}{MGCN \citep{mgcn}} & \multirow{3}{*}{Baby~\citep{amazon_reviews_2023}} & Original & 0.0944 & 0.0975 & 0.0960 & 0.0411 & 0.0430 & 0.0425 & 0.5912 & 0.0030 & 0.0934 & 0.4695 & 0.0008 & 0.0731 \\
\cmidrule(lr){3-15}
& & MMRecUn \citep{mmrecun} & 0.0883 & 0.0915 & 0.0893 & 0.0401 & 0.0411 & 0.0400 & \textbf{0} & 0.0002 & \textbf{0} & \textbf{0} & \textbf{0} & \textbf{0} \\
& & Ours & \cellcolor{RetainColor!15} \textbf{0.0944} & \cellcolor{RetainColor!14} \textbf{0.0974} & \cellcolor{RetainColor!17} \textbf{0.0966} & \cellcolor{RetainColor!8} \textbf{0.0411} & \cellcolor{RetainColor!13} \textbf{0.0429} & \cellcolor{RetainColor!18} \textbf{0.0426} & \textbf{0} & \textbf{0} & \textbf{0} & \textbf{0} & \textbf{0} & \textbf{0} \\
\cmidrule(lr){2-15}
& \multirow{3}{*}{Clothing~\citep{amazon_reviews_2023}} & Original & 0.0895 & 0.0918 & 0.0907 & 0.0406 & 0.0417 & 0.0412 & 0.8056 & 0.0035 & 0.1457 & 0.6441 & 0.0009 & 0.1158 \\
\cmidrule(lr){3-15}
& & MMRecUn \citep{mmrecun} & 0.0909 & 0.0887 & 0.0904 & 0.0401 & 0.0405 & 0.0407 & \textbf{0} & \textbf{0} & \textbf{0} & \textbf{0} & \textbf{0} & \textbf{0} \\
& & Ours & \cellcolor{RetainColor!9} \textbf{0.0945} & \cellcolor{RetainColor!14} \textbf{0.0942} & \textbf{0.0906} & \cellcolor{RetainColor!24} \textbf{0.0436} & \cellcolor{RetainColor!16} \textbf{0.0428} & \cellcolor{RetainColor!5} \textbf{0.0413} & \textbf{0} & \textbf{0} & \textbf{0} & \textbf{0} & \textbf{0} & \textbf{0} \\
\cmidrule(lr){2-15}
& \multirow{3}{*}{Sports~\citep{amazon_reviews_2023}} & Original & 0.1071 & 0.1093 & 0.1091 & 0.0471 & 0.0484 & 0.0484 & 0.4352 & 0.0002 & 0.0773 & 0.3186 & 0.0001 & 0.0554 \\
\cmidrule(lr){3-15}
& & MMRecUn \citep{mmrecun} & 0.1071 & 0.1101 & 0.1024 & 0.0484 & \textbf{0.0499} & 0.0465 & \textbf{0} & \textbf{0} & \textbf{0} & \textbf{0} & \textbf{0} & \textbf{0} \\
& & Ours & \cellcolor{RetainColor!3} \textbf{0.1080} & \textbf{0.1103} & \cellcolor{RetainColor!16} \textbf{0.1091} & \cellcolor{RetainColor!7} \textbf{0.0492} & 0.0493 & \cellcolor{RetainColor!14} \textbf{0.0485} & \textbf{0} & \textbf{0} & \textbf{0} & \textbf{0} & 0.0001 & \textbf{0} \\
\midrule
\multirow{9}{*}{MIG-GT \citep{hu2025modality}} & \multirow{3}{*}{Baby~\citep{amazon_reviews_2023}} & Original & 0.0987 & 0.0949 & 0.0968 & 0.0435 & 0.0409 & 0.0423 & 0.6787 & 0.6189 & 0.6158 & 0.5223 & 0.3235 & 0.3226 \\
\cmidrule(lr){3-15}
& & MMRecUn \citep{mmrecun} & 0.0825 & 0.0926 & 0.0816 & 0.0383 & 0.0410 & 0.0374 & 0.0775 & \textbf{0} & 0.0240 & 0.0515 & \textbf{0} & 0.0155 \\
& & Ours & \cellcolor{RetainColor!42} \textbf{0.1008} & \cellcolor{RetainColor!16} \textbf{0.0994} & \cellcolor{RetainColor!40} \textbf{0.0991} & \cellcolor{RetainColor!39} \textbf{0.0442} & \cellcolor{RetainColor!21} \textbf{0.0441} & \cellcolor{RetainColor!42} \textbf{0.0437} & \cellcolor{ForgetColor!12} \textbf{0.0132} & \textbf{0} & \cellcolor{ForgetColor!9} \textbf{0.0038} & \cellcolor{ForgetColor!11} \textbf{0.0095} & \textbf{0} & \cellcolor{ForgetColor!9} \textbf{0.0024} \\
\cmidrule(lr){2-15}
& \multirow{3}{*}{Clothing~\citep{amazon_reviews_2023}} & Original & 0.0903 & 0.0903 & 0.0889 & 0.0413 & 0.0409 & 0.0402 & 0.8383 & 0.7652 & 0.7917 & 0.6603 & 0.3804 & 0.4452 \\
\cmidrule(lr){3-15}
& & MMRecUn \citep{mmrecun} & 0.0847 & 0.0872 & 0.0820 & 0.0383 & 0.0393 & 0.0379 & 0.5234 & 0.0001 & 0.1457 & 0.4572 & \textbf{0} & 0.1025 \\
& & Ours & \cellcolor{RetainColor!17} \textbf{0.0917} & \cellcolor{RetainColor!9} \textbf{0.0905} & \cellcolor{RetainColor!16} \textbf{0.0885} & \cellcolor{RetainColor!23} \textbf{0.0417} & \cellcolor{RetainColor!14} \textbf{0.0412} & \cellcolor{RetainColor!19} \textbf{0.0406} & \cellcolor{ForgetColor!42} \textbf{0.0022} & \textbf{0} & \cellcolor{ForgetColor!11} \textbf{0.0976} & \cellcolor{ForgetColor!42} \textbf{0.0014} & \textbf{0} & \cellcolor{ForgetColor!11} \textbf{0.0877} \\
\cmidrule(lr){2-15}
& \multirow{3}{*}{Sports~\citep{amazon_reviews_2023}} & Original & 0.1092 & 0.1089 & 0.1100 & 0.0493 & 0.0492 & 0.0491 & 0.7510 & 0.6248 & 0.6502 & 0.6021 & 0.2939 & 0.3532 \\
\cmidrule(lr){3-15}
& & MMRecUn \citep{mmrecun} & \cellcolor{RetainColor!6}\textbf{0.1074} & \cellcolor{RetainColor!3}\textbf{0.1093} & 0.1015 & 0.0476 & 0.0489 & 0.0452 & 0.0457 & \textbf{0} & 0.1155 & 0.0301 & \textbf{0} & 0.0770 \\
& & Ours & 0.1036 & 0.1087 & \cellcolor{RetainColor!17} \textbf{0.1086} & \cellcolor{RetainColor!13} \textbf{0.0494} & \cellcolor{RetainColor!6} \textbf{0.0496} & \cellcolor{RetainColor!27} \textbf{0.0492} & \cellcolor{ForgetColor!11} \textbf{0.0051} & \textbf{0} & \cellcolor{ForgetColor!16} \textbf{0.0042} & \cellcolor{ForgetColor!10} \textbf{0.0034} & \textbf{0} & \cellcolor{ForgetColor!14} \textbf{0.0021} \\
\bottomrule
\end{tabular}
}
\vspace{-10pt}
\end{table*}

\textbf{Unified Update Mechanism and Algorithm Overview}
Integrating the above three strategies, the targeted reverse step for parameter group $k$ is
\begin{equation}
\theta_k
\leftarrow
\theta_k
-
\eta\,
z_k \, \gamma_{m(k)}
\nabla_{\theta_k}\mathcal{L}_{\mathrm{rev+gate}},
\label{eq:tru_reverse_step}
\end{equation}
where $\gamma_{m(k)}$ scales the reverse gradient for the branch containing group $k$, and $z_k$ masks unselected groups. As summarized in Algorithm~\ref{alg:tru}, each epoch first estimates and selects sensitive parameter groups, then estimates the branch scalers. TRU installs the corresponding gradient hooks for the reverse phase, removes them afterward, and finally repairs the model on retained batches.

\section{Experiments}\label{experiments}
\begin{figure}[t!]
  \centering
  \includegraphics[width=0.48\textwidth]{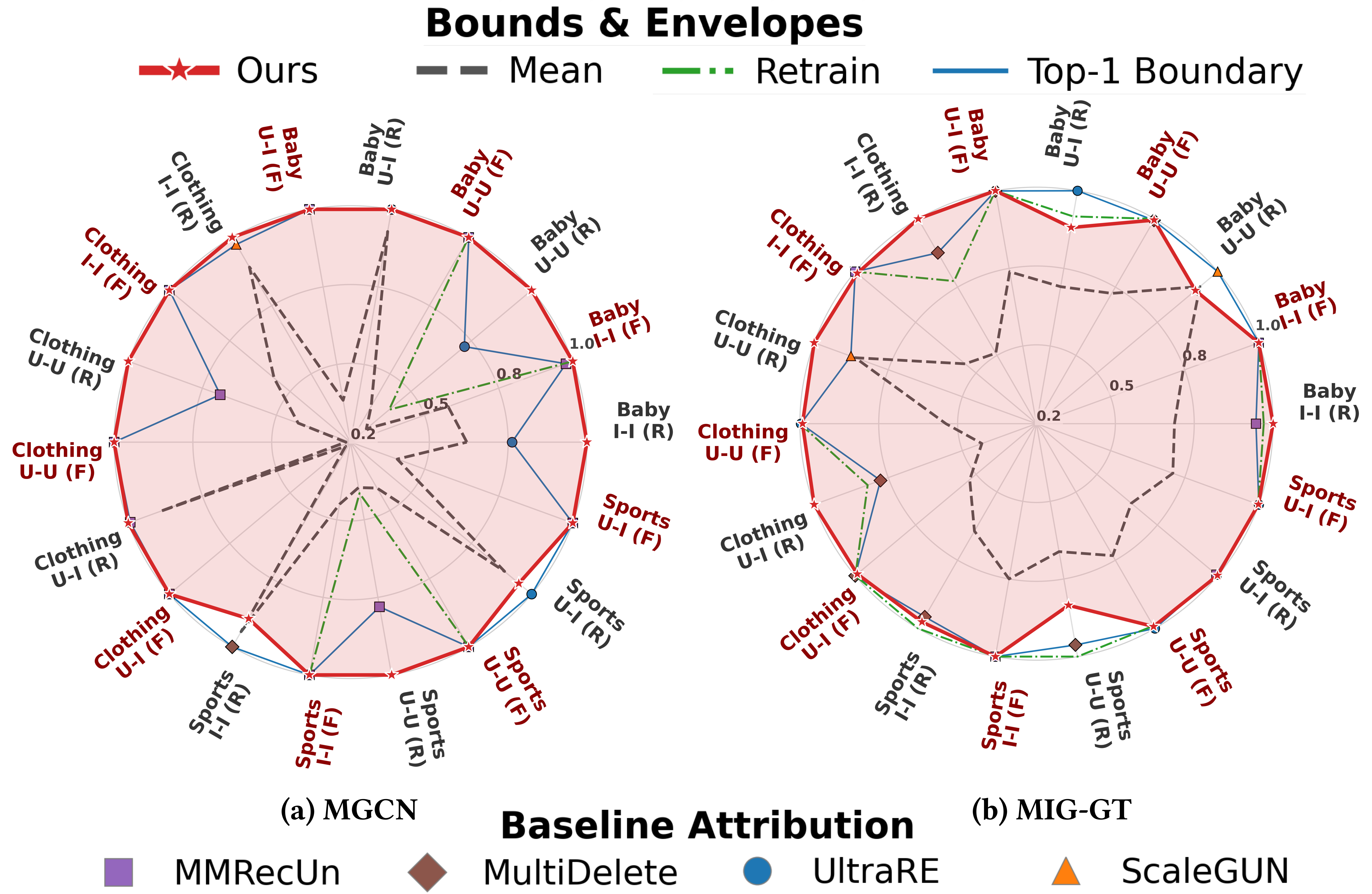}
  \vspace{-20pt}
  \caption{Normalized radar across all datasets, backbones, and unlearning regimes. TRU exhibits the most balanced overall profile between forget-side and retain-side objectives, suggesting that its advantage stems from superior trade-off control rather than isolated improvements under specific settings. Best viewed in color.}
  \Description{Normalized radar chart aggregating retain-side and forget-side objectives across datasets, backbones, and deletion regimes. TRU covers a broader and more balanced region than the compared approximate methods.}
  \label{fig:radar}
  \vspace{-10pt}
\end{figure}
In this section, we first introduce a fair and rigorous evaluation protocol for MRS unlearning and present a comparative analysis against baselines. Due to space constraints, comprehensive reproducibility details are provided in the Supplementary, including forget-side comparisons and hyperparameter sensitivity study.

\subsection{Experimental Setup}\label{sec:evaluation}
We compare six baselines: \emph{Original}, \emph{Retrain}, UltraRE~\citep{ultrare}, MultiDelete~\citep{multidelete}, ScaleGUN~\citep{scalegun_yi2024}, and MMRecUn~\citep{mmrecun}. Here, \emph{Original} denotes the model trained on the full training set, and \emph{Retrain} denotes a full retraining from scratch on the retained set. 
We use three distinct categories from the public Amazon review data~\citep{ni2019justifying,mcauley2015image,amazon_reviews_2023}, specifically \textbf{Baby}, \textbf{Sports}, and \textbf{Clothing}.

Following the CURE4Rec benchmark~\citep{benchmark24}, our evaluation framework balances three competing objectives: \textbf{forgetting completeness} (ensuring targeted data is thoroughly removed), \textbf{retained utility} (preserving the model's overall recommendation accuracy), and \textbf{efficiency} (the wall-clock time to achieve the behavior of a retrained model). To demonstrate real-world applicability, we systematically assess TRU across three distinct unlearning regimes: \textit{user-level} (simulating complete account deletion, denoted U-U; the main challenge scenario of MRS unlearning), \textit{item-level} (simulating global content takedowns, denoted I-I; this is \textbf{different} from \textit{Item-centric}, which we discussed in Section~\ref{sec:design}), and \textit{interaction-level} (allowing for granular user preference correction, denoted U-I).

To ensure a rigorous assessment of the inherent utility-forgetting trade-off, we \textbf{avoid single-metric evaluations} that might obscure degradation in model utility. Consequently, our protocol necessitates a dual-perspective approach, balancing \emph{FORGET} efficacy against \emph{RETAIN} performance:
\begin{itemize}[leftmargin=*]
    \item \textbf{Utility Balance:} We assess the Retain--Forget trade-off using standard ranking indicators~\citep{mmrecun} (\textit{e.g.}, Recall@20, NDCG@20). Recall@20-R evaluates whether retained positive interactions remain retrievable (higher is better), whereas Recall@20-F measures whether deleted positive pairs remain retrievable (lower is better). A successful approximate unlearning method must symmetrically approach the gold-standard performance frontier of exact retraining.
    \item \textbf{Security-Oriented Erasure:} Beyond surface-level utility, we audit structural forgetting via Membership Inference Attacks (MIA) \citep{miars} and Backdoor evaluations (BKD) \citep{backdoor_mu, benchmark24}. We prefer MIA Balanced Accuracy (\emph{BalAcc}) closer to $0.5$, indicating weaker membership leakage, and a post-unlearning Attack Success Rate (\emph{ASR}) closer to $0$, indicating more complete trigger removal.
\end{itemize}

\begin{table}[t]
\centering
\small
\caption{Illustrative user-level retain--forget trade-offs. Higher Recall@20-R and lower Recall@20-F are preferred. All ranking metrics are reported to four decimal places. Entries shown as 0 denote values below 0.00005 after rounding, not exact zero.}
\label{tab:user_setting_all_baselines_two_backbones}
\vspace{-0.15in}
\begin{tabular}{cl c c}
\toprule
\textbf{Backbone} & \textbf{Method} & \textbf{Recall@20-R} $\uparrow$ & \textbf{Recall@20-F} $\downarrow$ \\
\midrule
\multirow{7}{*}{\shortstack{MIG-GT~\citep{hu2025modality}\\(sports~\citep{amazon_reviews_2023})}} 
&  Original  & 0.1091 & 0.6834 \\
\cmidrule{2-4}
& Retrain           & 0.1046 \gup{0.0012} & 0.0095 \gdown{0.6723} \\
& UltraRE \citep{ultrare}        & 0.0780 \rdown{0.0254} & \textbf{0.0035} \gdown{0.6783} \\
& MultiDelete \citep{multidelete}    & 0.1019 \rdown{0.0015} & 0.0086 \gdown{0.6732} \\
& ScaleGUN \citep{scalegun_yi2024}       & \textbf{0.1085} \gup{0.0051} & 0.7303 \rup{0.0485} \\
& MMRecUn \citep{mmrecun}        & 0.1074 \gup{0.0040} & 0.0457 \gdown{0.6361} \\
&  \cellcolor{gray!15}\rule{0pt}{1.1em}\textbf{Ours}   &  \cellcolor{gray!15}\rule{0pt}{1.1em}0.1036 \gup{0.0002} &  \cellcolor{gray!15}\rule{0pt}{1.1em}0.0072 \gdown{0.6746} \\
\midrule
\multirow{7}{*}{\shortstack{MGCN~\citep{mgcn}\\(clothing~\citep{amazon_reviews_2023})}} 
& Original & 0.0929 & 0.8056 \\
\cmidrule{2-4}
& Retrain           & 0.0856 \rdown{0.0073} & 0.0045 \gdown{0.8011} \\
& UltraRE \citep{ultrare}        & 0.0829 \rdown{0.0100} & 0.8049 \gdown{0.0007} \\
& MultiDelete \citep{multidelete}    & 0.0892 \rdown{0.0037} & 0.8045 \gdown{0.0011} \\
& ScaleGUN \citep{scalegun_yi2024}       & 0.0884 \rdown{0.0045} & 0.8289 \rup{0.0233} \\
& MMRecUn \citep{mmrecun}        & 0.0909 \rdown{0.0020} & \textbf{0} \gdown{0.8056} \\
&  \cellcolor{gray!15}\rule{0pt}{1.1em}\textbf{Ours}   &  \cellcolor{gray!15}\rule{0pt}{1.1em}\textbf{0.0945} \gup{0.0016} &  \cellcolor{gray!15}\rule{0pt}{1.1em}\textbf{0} \gdown{0.8056} \\
\bottomrule
\end{tabular}
\vspace{-10pt}
\end{table}

\subsection{General Unlearning Performance}
\label{sec:exp_general}

Table~\ref{tab:comparison} and Figure~\ref{fig:radar} demonstrate that TRU yields a stronger retain–forget trade-off in most settings across both MGCN~\citep{mgcn} and MIG-GT~\citep{hu2025modality} backbones. We distill three key quantitative insights. 
\textbf{(1) TRU removes residual target signals more effectively.} Compared with prior approximate baselines, TRU pushes forget-side performance much closer to retraining, especially on harder settings where residual traces remain strong after uniform reverse updates. This trend is particularly evident on the more decoupled MIG-GT backbone, where baseline methods often leave non-trivial Forget Recall@20 after unlearning, whereas TRU leaves substantially less residual exposure.
\textbf{(2) TRU preserves retain-side utility.} When compared against the original model, TRU generally keeps Retain Recall@20 and Retain NDCG@20 much more stable than uniform-update baselines, indicating that its reverse updates are better localized and cause less collateral damage to useful retained knowledge. 
\textbf{(3) The advantage is consistent across deletion regimes.} This pattern is not confined to a single setting: across user-, item-, and interaction-level unlearning, TRU more reliably balances forgetting and retention, instead of improving one side by severely hurting the other. The radar plots in Figure~\ref{fig:radar} further visualize this point by showing that TRU expands the trade-off boundary more consistently toward the retraining frontier.

\begin{figure*}[t!]
  \centering
  \includegraphics[width=0.96\textwidth]{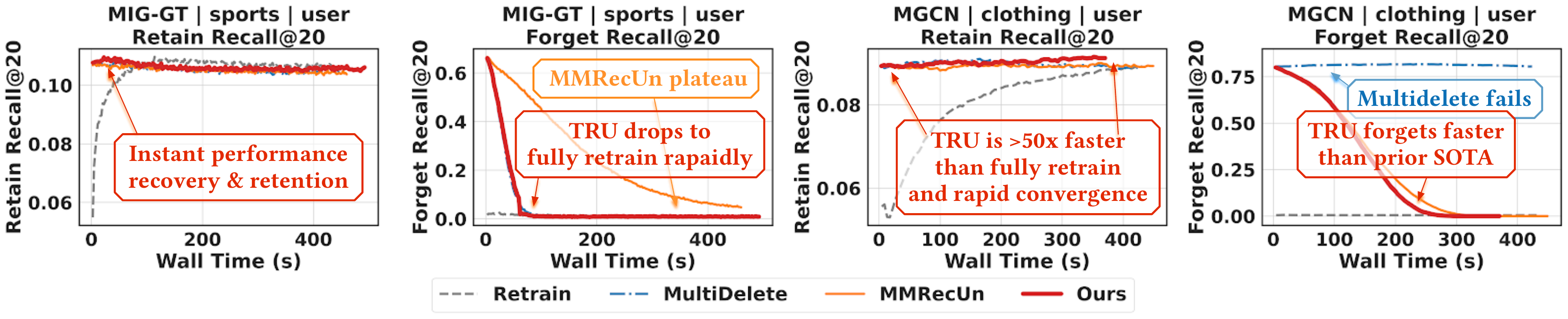}
  \vspace{-10pt}
 \caption{Wall-clock trajectories of retain-side and forget-side Recall@20 under the user-level unlearning. Read each pair left-to-right: retain utility (higher is better), then forget exposure (lower is better); the horizontal axis is wall-clock time.}
  \Description{Wall-clock curves for user-level unlearning, showing how retain-side and forget-side Recall at 20 evolve over time for TRU, retraining, and the compared unlearning baselines.}
  \label{fig:wall-time}
  \vspace{-10pt}
\end{figure*}



\begin{table}[t]
\centering
\small
\caption{Security-oriented erasure audit using membership inference attack (MIA) balanced accuracy and backdoor attack success rate (ASR). For MIA, values closer to 0.5 indicate weaker membership leakage; for BKD, lower ASR is better. Retrain is reported as the exact-unlearning reference to contextualize approximate methods.}
\vspace{-0.15in}
\label{tab:mia+bkd}
\begin{tabular}{cl c c}
\toprule
\textbf{Backbone} & \textbf{Method} & \textbf{BalAcc} $\rightarrow 0.5$ &  \textbf{ASR} $\rightarrow 0.0$ \\
\midrule
\multirow{6}{*}{\shortstack{MIG-GT~\citep{hu2025modality}\\(sports~\citep{amazon_reviews_2023})}} 
& Retrain & 0.6153 & 0.1622 \\
\cmidrule{2-4}
& UltraRE~\citep{ultrare}          & 0.6044 \gdown{0.0109} & 1.0000 \rup{0.8378} \\
& MultiDelete~\citep{multidelete}  & 0.5686 \gdown{0.0467} & \textbf{0.0811} \gdown{0.0811} \\
& ScaleGUN~\citep{scalegun_yi2024} & 0.5786 \gdown{0.0367} & 0.2432 \rup{0.0810} \\
& MMRecUn~\citep{mmrecun}            & 0.5626 \gdown{0.0527} & 0.2162 \rup{0.0540} \\
&  \cellcolor{gray!15}\rule{0pt}{1.1em}\textbf{Ours} &  \cellcolor{gray!15}\rule{0pt}{1.1em}\textbf{0.5253} \gdown{0.0900} &  \cellcolor{gray!15}\rule{0pt}{1.1em}\textbf{0.0811} \gdown{0.0811} \\
\midrule
\multirow{6}{*}{\shortstack{MGCN~\citep{mgcn}\\(clothing~\citep{amazon_reviews_2023})}} 
& Retrain & 0.5156 & 0 \\
\cmidrule{2-4}
& UltraRE~\citep{ultrare}         & 0.9440 \rup{0.4284} & 0.1333 \rup{0.1333} \\
& MultiDelete~\citep{multidelete}  & 0.9452 \rup{0.4296} & 0.1333 \rup{0.1333} \\
& ScaleGUN~\citep{scalegun_yi2024} & 0.9442 \rup{0.4286} & 0.1333 \rup{0.1333} \\
& MMRecUn~\citep{mmrecun}         & 0.9430 \rup{0.4274} & 0.1333 \rup{0.1333} \\
&  \cellcolor{gray!15}\rule{0pt}{1.1em}\textbf{Ours} &  \cellcolor{gray!15}\rule{0pt}{1.1em}\textbf{0.9388} \rup{0.4232} &  \cellcolor{gray!15}\rule{0pt}{1.1em}0.1333 \rup{0.1333} \\
\bottomrule
\end{tabular}
\vspace{-10pt}
\end{table}

\subsection{In-Depth Analysis on Challenging Scenarios}
\label{sec:exp_specific}

While aggregate results summarize overall performance, we further examine two challenging user-level settings: MIG-GT~\citep{hu2025modality} on Sports~\citep{amazon_reviews_2023} and MGCN~\citep{mgcn} on Clothing~\citep{amazon_reviews_2023}. These settings expose distinct retain--forget trade-offs, which we analyze across utility, security, and wall-clock efficiency.

\textbf{(1) Retain--Forget Trade-offs:} Table~\ref{tab:user_setting_all_baselines_two_backbones} illustrates distinct retain--forget trade-offs. On MIG-GT~\citep{hu2025modality} (Sports~\citep{amazon_reviews_2023}), TRU reaches $0.1036$ Retain Recall and $0.0072$ Forget Recall, remaining close to the retraining reference ($0.1046$ and $0.0095$) while leaving less residual exposure than several approximate baselines at comparable retain utility. On MGCN~\citep{mgcn} (Clothing~\citep{amazon_reviews_2023}), TRU achieves higher Retain Recall ($0.0945$ vs.\ $0.0909$ for MMRecUn~\citep{mmrecun}) and matches the best rounded Forget Recall: both values are below $0.00005$ (displayed as 0 after four-decimal rounding).

\textbf{(2) Auditing Adversarial Traces:} Surface-level ranking metrics alone do not guarantee that deleted data becomes unrecoverable, so Table~\ref{tab:mia+bkd} further audits privacy leakage and attackability using MIA and backdoor metrics. On MIG-GT~\citep{hu2025modality} (Sports~\citep{amazon_reviews_2023}), TRU obtains the lowest MIA BalAcc among the approximate methods ($0.5253$, closest to the ideal $0.5$) and ties for the lowest ASR ($0.0811$). The Retrain reference in this setting is $0.6153$ BalAcc and $0.1622$ ASR. On MGCN~\citep{mgcn} (Clothing~\citep{amazon_reviews_2023}), TRU has the lowest approximate BalAcc ($0.9388$), while all approximate methods remain far from $0.5$ and report the same ASR of $0.1333$. The result therefore indicates partial mitigation in this setting.

\textbf{(3) Accelerating Practical Convergence:} Figure~\ref{fig:wall-time} demonstrates that TRU translates its targeted interventions into strictly superior computational efficiency. Prior methods like MMRecUn~\citep{mmrecun} expend substantial computational budgets blindly perturbing weakly relevant interactions and insensitive layers, leading to sluggish and unstable optimization. By structurally isolating the reverse updates—via fusion gating, modality scaling, and capacity-aware masking—TRU entirely eliminates this computational waste. Consequently, on MIG-GT~\citep{hu2025modality} (Sports~\citep{amazon_reviews_2023}), TRU rapidly suppresses Forget Recall while stabilizing Retain utility in a high-value band significantly earlier than all competitors, achieving the optimal operating region at a fraction of the exact retraining latency.

\subsection{Ablation Studies}
\label{sec:exp_ablation}


\begin{table}[t]
\centering
\small
\caption{Component ablation on the Baby dataset under the user-level setting, focusing on forget-side effectiveness. All variants already reduce ASR$_{\text{after}}$ to zero in the BKD audit, so the remaining BKD difference reflects the residual poisoned influence on clean inputs after unlearning. All ranking metrics are reported to four decimal places. Entries shown as 0 denote values below 0.00005 after rounding, not exact zero.}
\label{tab:ablation}
\vspace{-0.15in}
\begin{tabular}{cl c c}
\toprule
\textbf{Backbone} & \textbf{Ablation} & \textbf{Recall@20-F} $\downarrow$ & \textbf{Clean$_{\text{after}}$} $\downarrow$ \\
\midrule
\multirow{4}{*}[-0.5ex]{\shortstack[l]{MGCN~\citep{mgcn}}} 
& \cellcolor{gray!15}\rule{0pt}{1.1em}Full Model & \cellcolor{gray!15}\rule{0pt}{1.1em}\textbf{0} & \cellcolor{gray!15}\rule{0pt}{1.1em}\textbf{0.0871} \\
\cmidrule{2-4}
& w/o Ranking Gating   & 0.0746\rup{0.0746} & 0.0904\rup{0.0033} \\
& w/o Modality Scaling & 0.0074\rup{0.0074} & 0.0903\rup{0.0032} \\
& w/o Group Selection  & 0.0075\rup{0.0075} & 0.0906\rup{0.0035} \\
\midrule
\multirow{4}{*}[-0.5ex]{\shortstack[l]{MIG-GT~\citep{hu2025modality}}} 
& \cellcolor{gray!15}\rule{0pt}{1.1em}Full Model & \cellcolor{gray!15}\rule{0pt}{1.1em}\textbf{0.0077} & \cellcolor{gray!15}\rule{0pt}{1.1em}\textbf{0.0934} \\
\cmidrule{2-4}
& w/o Ranking Gate   & 0.0153\rup{0.0076} & 0.0953\rup{0.0019} \\
& w/o Modality Scaling & 0.0081\rup{0.0004} & 0.0950\rup{0.0016} \\
& w/o Group Selection  & 0.0093\rup{0.0016} & 0.0959\rup{0.0025} \\
\bottomrule
\end{tabular}
\vspace{-10pt}
\end{table}

To validate the structural necessity of TRU's individual components, Table~\ref{tab:ablation} deconstructs the framework under the user-level deletion setting, the main challenge scenario of MRS unlearning. 
\textbf{(1) Ranking Gate Governs Surface Exposure:} Removing the fusion gate triggers the most severe failure in output-side erasure. On MGCN~\citep{mgcn}, the Forget Recall@20 drastically spikes from $0$ in the full model to $0.0746$ without the gate, confirming that actively sparsifying cross-modal fusion is the primary defense against collaborative popularity inertia. 
\textbf{(2) Parameter-Group Selection Eradicates Deep Residuals:} While all ablated variants successfully reduce explicit backdoor triggers to zero, auditing the clean-data outputs (\emph{Clean$_{\text{after}}$}) reveals deeper latent traces. Removing the capacity-aware module mask causes the clearest spike in this deep residual across both backbones (\textit{e.g.}, $0.0934 \rightarrow 0.0959$ on MIG-GT~\citep{hu2025modality}). This quantitatively proves that unlearning remains superficial unless reverse updates are strictly isolated to structurally sensitive modules within the backbone.
\textbf{(3) Modality Scaling Provides Guardrails:} Omitting branch-wise scaling yields strictly consistent degradations across both surface utility and deep security views. On MGCN~\citep{mgcn}, its removal simultaneously worsens Forget Recall@20 ($0 \rightarrow 0.0074$) and the clean residual trace ($0.0871 \rightarrow 0.0903$), substantiating that decoupled multimodal branches intrinsically cannot absorb uniform gradients without risking incomplete deletion. Together, these metrics validate that TRU's three components act as an indispensable, strictly complementary triad.

\section{Conclusion}

In this paper, we study approximate unlearning in multimodal recommendation systems. The main difficulty is not simply applying a stronger reverse update, but removing deleted-data influence without damaging shared representations needed by retained users and items. Our analysis identifies three bottlenecks in MRS: item-side residual traces, uneven modality responses, and concentrated parameter sensitivity. Based on these observations, we propose TRU, which refines the reverse pass through Ranking Gate, Modality Scaling, and Parameter-Group Selection. The gate attenuates residual target-item exposure, modality scaling calibrates branch-wise gradients, and group selection confines reverse updates to deletion-sensitive modules. This coordinated design targets the three bottlenecks without applying the same perturbation to every model component. Across the reported settings, TRU generally improves the trade-off between forgetting and retained utility over the evaluated approximate baselines. Ranking, security, and wall-clock evaluations provide complementary evidence of retained utility, residual exposure, and deletion-trace detectability. Together, these results suggest that practical MRS unlearning benefits from joint control at the output, representation, and parameter levels.

Future work will extend targeted unlearning beyond efficient deletion toward a broader responsible multimedia setting. 
In particular, it is promising to study how MRS can support auditable and continuously deployable unlearning under evolving data, model updates, and richer modalities, while jointly considering privacy, transparency, fairness, and robustness. 
Such deployment also requires stable deletion behavior when requests arrive sequentially and catalog content changes. Evaluating these conditions would connect the present one-shot audits to longer-term operational guarantees for responsible recommendation services.

\clearpage
\begin{acks}
This work was supported by the Fundamental and Interdisciplinary Disciplines Breakthrough Plan of the Ministry of Education of China (JYB2025XDXM102), the Sichuan Province Innovative Talent Funding Project for Postdoctoral Fellows (BX202405), and the Sichuan Science and Technology Program (2026NSFSC1451).
\end{acks}

\bibliographystyle{ACM-Reference-Format}
\balance
\bibliography{sample-base}

\clearpage
\appendix

\section{Detailed Experiment Setup} \label{app:detailed}

\subsection{Setup.}\label{setup}
We conducted experiments on a single NVIDIA GeForce RTX 4080 GPU with CUDA 12.1 and Python 3.10. Unless otherwise stated, we used Adam~\citep{adam} with learning rate $\eta=0.001$ and batch size $b=2048$. We used a unified experimental wrapper for all methods, consisting of base training, method-specific unlearning, an optional retain-side repair stage, and final evaluation.

\textbf{Baselines.}
We compare six systems: \emph{Original}, \emph{Retrain}, UltraRE~\citep{ultrare}, MultiDelete~\citep{multidelete}, ScaleGUN~\citep{scalegun_yi2024}, and MMRecUn~\citep{mmrecun}. Here, \emph{Original} denotes the model trained on the full training set, and \emph{Retrain} denotes exact retraining from scratch on the retain set. For approximate baselines, we evaluate them under the same outer experimental wrapper. When the repair stage is enabled, we apply an additional retain-only repair step after unlearning. This repair step is part of our unified evaluation protocol and should not be interpreted as a native component of every baseline.

\textbf{Datasets.}
We use three Amazon categories widely adopted in multimodal recommendation benchmarks, namely Baby, Sports, and Clothing~\citep{amazon,ni2019justifying}. For each dataset, we split the interactions into training, validation, and test sets with a ratio of 8:1:1. The forget set and the retain set are both constructed from the original training split, so that all unlearning requests are issued only on training interactions.

\textbf{Hyper-parameters.}
We tune three hyper-parameters in a small grid:
\textbf{(i)} the \textbf{top-ratio} $r \in \{0.1, 0.2, \ldots, 0.9\}$ for Layer Selection, where $r$ is exactly the same quantity as the top proportion $p$ in the main text; it controls the proportion of top-sensitive layers retained by the adaptive selection operator (denoted as $\mathrm{AdaptiveTop}(\cdot)$ in Eq.~\eqref{eq:layer_mask});
\textbf{(ii)} the \textbf{mini-batch budget} $M \in \{1,2,3,4,5\}$ used to estimate layer sensitivity and modality statistics at the beginning of each epoch, which is the same per-epoch statistics budget used in the implementation;
and
\textbf{(iii)} whether \textbf{Modality Scaling} is enabled (\texttt{retain-scale} $\in \{\text{on}, \text{off}\}$), corresponding to whether branch-wise retain-guided scaling is activated during the reverse step.
When \texttt{retain-scale} is \textit{on}, the reverse step uses branch-wise scaling factors as in Eq.~\eqref{eq:modality_scale}. In addition, the minimum-capacity constraint in the main text is controlled by $\tau_{\min}$ in Eq.~\eqref{eq:layer_mask}, which is kept fixed here rather than tuned in this grid. All other settings remain fixed.

\subsection{Detailed Setup}

\begin{table*}[t]
  \centering
  \caption{Three Amazon multimodal recommendation datasets: overall and split-level statistics }
  \label{tab:amazon_multimodal_stats}
  \begin{tabular}{llrrrrrc}
    \toprule
    Dataset & Split     & \#Users & \#Items & \#Interactions & Avg act/user & Avg act/item & Sparsity (\%) \\
    \midrule
    \multirow{4}{*}{Baby} &
      Overall   & 19\,445 & 7\,050  & 160\,792 & 8.2691 & 22.8074 & 99.8827 \\
    & Training  & 19\,445 & 7\,047  & 118\,551 & 6.0967 & 16.8229 & 99.9135 \\
    & Validation& 19\,445 & 5\,483  & 20\,559  & 1.0573 & 3.7496  & 99.9807 \\
    & Testing   & 19\,445 & 5\,549  & 21\,682  & 1.1150 & 3.9074  & 99.9799 \\
    \midrule
    \multirow{4}{*}{Sports} &
      Overall   & 35\,598 & 18\,357 & 296\,337 & 8.3245 & 16.1430 & 99.9547 \\
    & Training  & 35\,598 & 18\,352 & 218\,409 & 6.1354 & 11.9011 & 99.9666 \\
    & Validation& 35\,598 & 13\,342 & 37\,899  & 1.0646 & 2.8406  & 99.9920 \\
    & Testing   & 35\,598 & 13\,738 & 40\,029  & 1.1245 & 2.9137  & 99.9918 \\
    \midrule
    \multirow{4}{*}{Clothing} &
      Overall   & 39\,387 & 23\,033 & 278\,677 & 7.0754 & 12.0990 & 99.9693 \\
    & Training  & 39\,387 & 23\,020 & 197\,338 & 5.0102 & 8.5725  & 99.9782 \\
    & Validation& 39\,387 & 16\,702 & 40\,150  & 1.0194 & 2.4039  & 99.9939 \\
    & Testing   & 39\,387 & 16\,803 & 41\,189  & 1.0458 & 2.4513  & 99.9938 \\
    \bottomrule
  \end{tabular}
  \vspace{4pt}
\end{table*}

\subsubsection{Backbones and Model-Specific Setups}
\label{sec:setup-models}

\paragraph{Backbones considered.}
We evaluate two multimodal graph-based recommenders as backbones:
(1) \textbf{MIG-GT}: Modality-Independent GNNs with Global Transformers~\citep{hu2025modality}, which decouples per-modality receptive fields and augments global context via a sampling-based Transformer;
(2) \textbf{MGCN}: a multimodal graph-based recommender in the MMGCN family~\citep{mgcn}, which fuses ID-, image-, and text-side signals on the user-item graph.

\paragraph{Common protocol.}
Unless otherwise stated, we unify the embedding size to $d=64$, optimize with BPR loss and $\ell_2$ regularization, and adopt full-ranking evaluation. All models are evaluated under the same data split, the same candidate pool construction, and the same top-$K$ recommendation protocol.

\subsubsection{Baselines and Detailed Experimental Setup}

\paragraph{MMRecUn (AAAI'25).}
MMRecUn is the most direct prior baseline tailored to multimodal recommendation unlearning. It performs a reverse-style update on the forget set together with a retain-side repair objective~\citep{mmrecun}. In our experiments, we set the mixing factor to $\alpha=0.2$. We evaluate MMRecUn under the same data split and full-ranking protocol as the other baselines.

\paragraph{UltraRE (NeurIPS'23).}
UltraRE enhances RecEraser \citep{recu} through an error-decomposition view~\citep{ultrare}. We follow its three-stage design: (i) partition the training set into $k$ shards; (ii) train one shard model per partition; and (iii) learn a validation-based combiner $\beta$ for aggregation. In our multimodal recommendation setting, each shard model uses the same MGCN backbone, while the combiner operates on the resulting fused user/item representations. We set $k{=}10$, Sinkhorn regularization $\varepsilon{=}0.1$, clustering outer iterations to $10$, sub-model training epochs to $3$, and the combiner learning rate to $10^{-2}$ with weight decay $0$.

\paragraph{MultiDelete (ECCV'24).}
MultiDelete is a multimodal machine unlearning method that decouples cross-modal associations while preserving unimodal competence~\citep{multidelete}. We instantiate its loss on top of MGCN by combining decoupling, multimodal knowledge retention, and unimodal knowledge retention terms. We set $\lambda_{\mathrm{dec}}=\lambda_{\mathrm{mkr}}=\lambda_{\mathrm{ukr}}=1.0$ and use \texttt{readout=concat} to match the fused representation used in our recommender.

\paragraph{ScaleGUN (adapted baseline).}
ScaleGUN is originally a certified graph unlearning method designed for graph settings such as edge, node, and node-feature unlearning~\citep{scalegun_yi2024}. In our experiments, we use an edge-style adaptation of ScaleGUN within the same recommendation-oriented evaluation wrapper. Therefore, the reported results should be interpreted as an adapted baseline under a unified multimodal recommendation protocol, rather than a theorem-preserving reproduction of the original certified graph setting.

\paragraph{Why these multimodal adjustments?}
Some baselines were not originally proposed for multimodal recommendation. To make them comparable in our setting, we adapt them at the representation and evaluation levels while preserving their original design intent as much as possible. Specifically, UltraRE is instantiated on multimodal user/item representations produced by the recommender backbone; MultiDelete is applied directly to fused and per-modality streams; and ScaleGUN is adapted to the edge-style unlearning setting used in our recommendation pipeline.

\paragraph{Evaluation protocol.}
We evaluate both recommendation utility and unlearning effectiveness under the same full-ranking protocol. Standard validation and test metrics are computed on the original validation and test splits. For forget-side and retain-side audits, we additionally build evaluation loaders directly from the corresponding $(u,i)$ pairs in the forget and retain sets. In these two subset evaluations, the positive pairs are not masked again, because they are exactly the targets to be audited. Unless otherwise stated, lower scores on the forget set indicate stronger forgetting, while higher scores on the validation, test, and retain sets indicate better utility preservation.

\begin{figure}[t]
  \centering
  \includegraphics[width=0.48\textwidth]{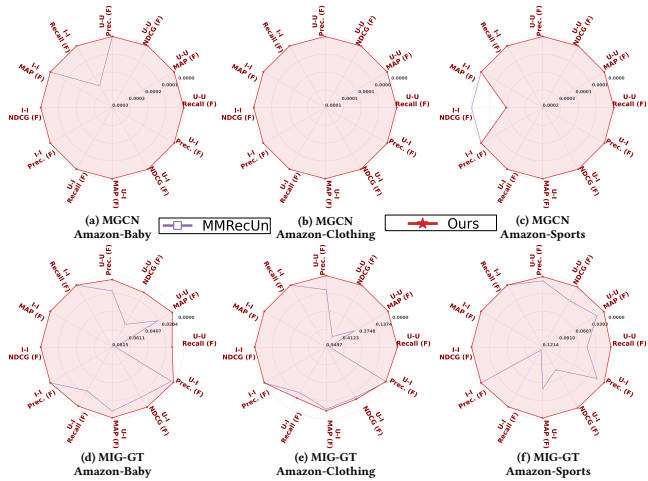}
  \vspace{-20pt}
  \caption{Normalized forget-side comparison between TRU and MMRecUn~\citep{mmrecun} across all settings. TRU is consistently competitive and generally stronger on the reported forget-side metrics, indicating that its advantage is not confined to a single dataset or backbone. Zoom in for details.}
  \label{fig:various_radar}
  \vspace{-10pt}
\end{figure}

\begin{figure}[t]
  \centering
  \includegraphics[width=0.48\textwidth]{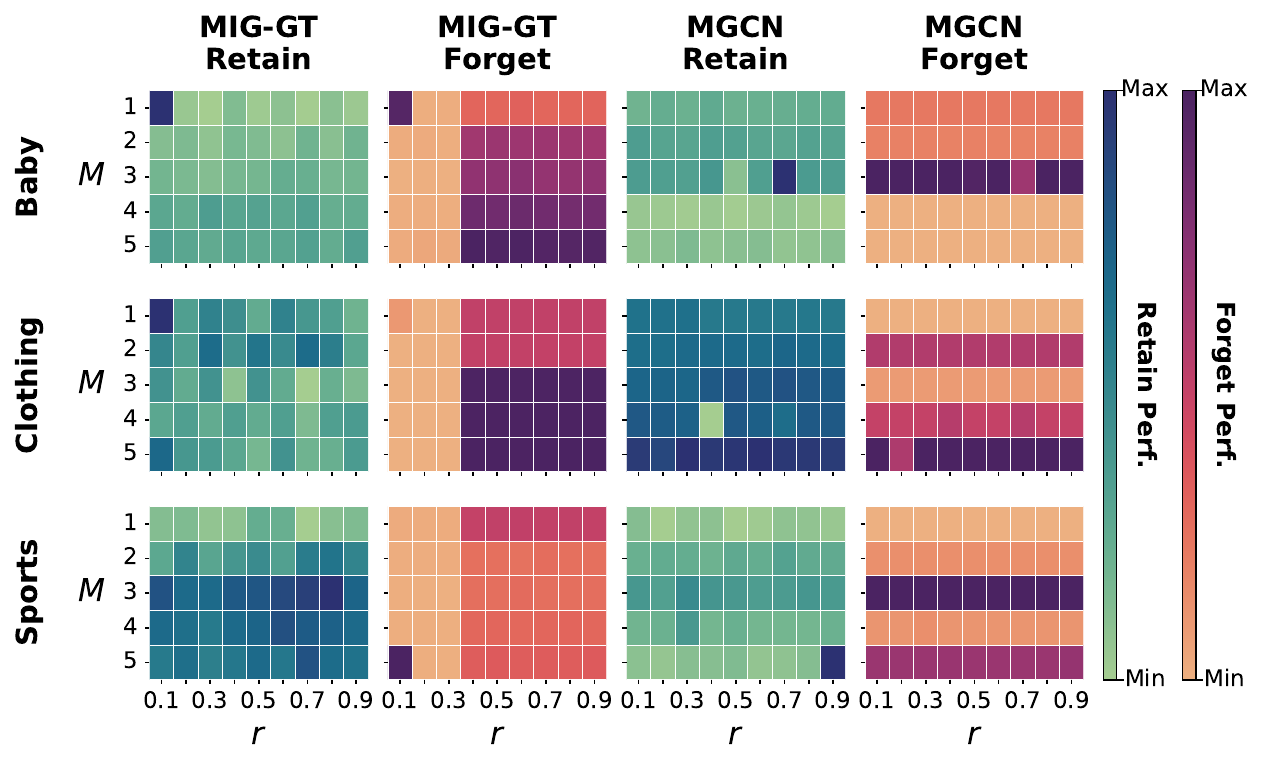}
  \vspace{-20pt}
  \caption{Hyper-parameter sensitivity of TRU over the tuned controls $(p, M)$ with modality scaling enabled. Across datasets and backbones, strong forget-side and retain-side performance appears in broad contiguous regions rather than isolated optima, indicating stable tuning behavior in practice.}
  \label{fig:hyper}
  \vspace{-10pt}
\end{figure}

\paragraph{Forgetting regimes.}
We instantiate three granularities of forgetting on the training interactions.
Let $\mathcal{D}_{\text{train}}$ be the training set and $\mathcal{U},\mathcal{I}$ the user and item universes.

\emph{(i) User-level forgetting (U-F).}
We sample a subset of users $\mathcal{U}_f \subset \mathcal{U}$ and define
$\mathcal{D}_f^{\text{U}}=\{(u,i)\in\mathcal{D}_{\text{train}}: u\in\mathcal{U}_f\}$.
The retain set is $\mathcal{D}_r=\mathcal{D}_{\text{train}}\setminus \mathcal{D}_f^{\text{U}}$.

\emph{(ii) Item-level forgetting (I-F).}
We sample a subset of items $\mathcal{I}_f \subset \mathcal{I}$ and define
$\mathcal{D}_f^{\text{I}}=\{(u,i)\in\mathcal{D}_{\text{train}}: i\in\mathcal{I}_f\}$,
with $\mathcal{D}_r=\mathcal{D}_{\text{train}}\setminus \mathcal{D}_f^{\text{I}}$.

\emph{(iii) Interaction-level forgetting (UI-F).}
We directly sample a subset of interactions
$\mathcal{D}_f^{\text{UI}}\subset \mathcal{D}_{\text{train}}$
and define $\mathcal{D}_r=\mathcal{D}_{\text{train}}\setminus \mathcal{D}_f^{\text{UI}}$.

\textbf{Protocols: Original / Retrain / Unlearned.}
For each regime, we report three kinds of systems:
(1) \emph{Original}, trained on $\mathcal{D}_{\text{train}}$;
(2) \emph{Retrain}, trained from scratch on $\mathcal{D}_r$;
(3) \emph{Unlearned}, obtained by applying the corresponding unlearning method to the original model with access to $(\mathcal{D}_f,\mathcal{D}_r)$.

\subsection{Additional Experimental Details}

\subsubsection{Comparison with MMRecUn}

\textbf{TRU consistently surpasses the most direct prior MMRS unlearning baseline.}
Figure~\ref{fig:various_radar} provides a normalized forget-side comparison between TRU and MMRecUn~\citep{mmrecun} across all settings.
The key observation is not merely that TRU is competitive on average, but that it is \emph{consistently stronger on all four metrics}.
This result is important because MMRecUn is the most direct prior approximate method specifically introduced for multimodal recommendation unlearning, rather than a generic unlearning or recommendation baseline.
Therefore, the advantage shown in Figure~\ref{fig:various_radar} indicates that TRU improves not only over broad baselines in the main tables, but also over the most relevant prior method designed for the same problem setting.

More importantly, the superiority is \emph{uniform rather than selective}.
TRU does not win by improving one metric while sacrificing the others.
Instead, it remains better across the entire four-metric forget-side profile after normalization.
This makes the comparison substantially stronger: the gain of TRU is not tied to one particular view of forgetting, but persists under multiple complementary criteria.

\subsubsection{Hyper-Parameter Sensitivity}\label{sec:hyper}

\textbf{TRU is stable and easy to tune in practice.}
Figure~\ref{fig:hyper} visualizes the performance landscape of TRU over the top proportion $p$ and mini-batch budget $M$.
A clear and practically useful pattern emerges: favorable forget-side and retain-side results occupy \emph{broad contiguous regions} instead of appearing only at a few isolated points.
This means that TRU does not rely on brittle hyper-parameter coincidence.
Its good performance can be reached by a relatively wide range of configurations, which makes parameter selection much easier in real use.

This pattern also makes TRU more reproducible.
When a method only works at a few scattered optima, tuning becomes expensive and unstable across datasets or backbones.
In contrast, the landscapes in Figure~\ref{fig:hyper} show visually coherent high-performing zones, indicating that TRU admits a structured operating region rather than a narrow sweet spot.
From a practical perspective, this is exactly the kind of sensitivity pattern that is easier to deploy, easier to transfer, and less likely to fail under modest tuning deviations.

Finally, the optimal regions are not identical across MGCN and MIG-GT.
This difference is itself informative: it shows that $p$ and $M$ act as meaningful control variables that adapt to backbone-specific characteristics, rather than redundant knobs with little functional effect.

\end{document}